\documentclass{svproc}

\usepackage{type1cm}        % activate if the above 3 fonts are
\usepackage{makeidx}         % allows index generation
\usepackage{graphicx}        % standard LaTeX graphics tool
\usepackage{multicol}        % used for the two-column index
\usepackage{multirow}% http://ctan.org/pkg/multirow
\usepackage[bottom]{footmisc}% places footnotes at page bottom
\usepackage{booktabs}

\usepackage[varvw]{newtxmath}       % selects Times Roman as basic font
\makeindex             % used for the subject index
\newcommand{\real}{\mathbb{R}}

\newcommand{\GG}{\mathcal{G}}
\newcommand{\VV}{\mathcal{V}}
\newcommand{\EE}{\mathcal{E}}
\newcommand{\NN}{\mathcal{N}}
\newcommand{\A}{\mathcal{A}}
\newcommand{\Scal}{\mathcal{S}}
\newcommand{\FF}{\mathcal{F}}
\newcommand{\BB}{\mathcal{B}}

\newtheorem{assumption}{Assumption}

\newcommand{\xyudel}[1]{}

\usepackage{todonotes}
\usepackage[normalem]{ulem}

\usepackage[]{algorithm2e}
\usepackage{bm}
\usepackage{color}
\graphicspath{ {images/} }
\usepackage{subcaption}

\begin{document}
\mainmatter              % start of a contribution
\title{
Receding Horizon Control on the Broadcast
of Information
in Stochastic Networks
}

\titlerunning{Receding Horizon Control on the Broadcast
of Information}

\author{Thales C. Silva\inst{1} \and Li Shen\inst{1} \and Xi Yu\inst{2} \and M. Ani Hsieh\inst{1}
% \thanks{This work was supported in part by the ARL Grant DCIST CRA W911NF-17-2-0181 and in part by the ONR Award No. N00014-17-1-2690.}
 }
 
\authorrunning{Thales C. Silva et al.} % abbreviated author list (for running head)
\tocauthor{Thales C. Silva, Li Shen, Xi Yu, and M. Ani Hsieh}

\institute{University of Pennsylvania,
Department of Mechanical Engineering and Applied Mechanics, 
Philadelphia, PA 19104, USA\\
\email{scthales@seas.upenn.edu}
\and
West Virginia University,
Department of Mechanical and Aerospace Engineering,\\
Morgantown, WV 26501, USA\\
\email{xi.yu1@mail.wvu.edu}
}

%	{\small Department of Mechanical Engineering and Applied Mechanics} \\
%	University of Pennsylvania, Philadelphia, USA \\
%		\{xyureka,m.hsieh\}@seas.upenn.edu
%\thanks{*We gratefully acknowledge the support of ONR Award No. N00014-17-1-2690 and ARL DCIST CRA W911NF-17-2-0181.}

\maketitle
%\thispagestyle{empty}
%\pagestyle{empty}

%%%%%%%%%%%%%%%%%%%%%%%%%%%%%%%%%%%%%%%%%%%%%%%%%%%%%%%%%%%%%%%%%%%%%%%%%%%%%%%%

\begin{abstract}
This paper focuses on the broadcast of information on
robot networks with stochastic network interconnection topologies.
Problematic communication networks are almost
unavoidable
in areas where we wish to deploy multi-robotic systems,
usually due to a lack of environmental consistency,
accessibility, and structure.
We tackle this problem by modeling the broadcast
of information in a multi-robot communication network as a stochastic process with random arrival
times, which can be produced by irregular robot movements,
wireless attenuation, and other environmental factors.
Using this model, we provide and analyze a receding horizon control
strategy to control the statistics of the information broadcast.
The resulting strategy compels the robots to re-direct their
communication resources to different neighbors according to
the current propagation process 
to fulfill global broadcast
requirements. Based on this method, we provide an approach to compute the expected time to broadcast the
message to all nodes. 
Numerical examples are provided to
illustrate the results.
\keywords{multi-robot system, receding horizon control, broadcast of information}
\end{abstract}

\section{Introduction}
\label{intro}

We are interested in leveraging spatio-temporal sensing capabilities of robotic teams to explore or monitor large-scale complex environments
(\textit{e.g.,} forests, oceans, and underground caves and tunnels \cite{breitenmoser2010voronoi,cassandras2016smart,gusrialdi2008voronoi,wei2018,yu2019synchronous}).
%\xyucomment{I mean 'deliver the tasks'. I edit the sentence accordingly. Thanks!}
While individual robots can each monitor or explore limited regions, a team of them can efficiently cover larger areas.  Naturally, individual robots in swarms are expected
to exchange information and make certain decisions based on the information gathered by itself and received from other robots \cite{hsieh2008decentralized,moarref2020automated}.
It is, therefore, essential for any robot to be able to
transmit its gathered data to others in the team. 

As the demand of information transmission can emerge between any pair of nodes, we use a propagation model to study the efficiency of the connectivity of the whole team. 
In such models, robots are represented as nodes of a graph,
and interactions 
%(\textit{i.e.} communication)
between nodes are represented by edges.
%The nodes and the edges together form a network.
At the beginning of a propagation, the nodes 
%(\textit{i.e.} robots) 
are considered as `non-informed', and will be `informed’ by the information propagated through the graph. A faster convergence of all nodes into the status of `informed' indicates a more \textit{efficiently connected} network. 
Existing works abstract the propagation of information from any individual agent to the rest of the team as compartmental models \cite{huang2006information,khelil2002epidemic,zanette2002dynamics}, which are also broadly used in modeling the spread of an infectious diseases \cite{brauer2008compartmental}.
%In such models, robots are represented as nodes of a graph,
%and interactions 
%(\textit{i.e.} communication)
%between nodes are represented by edges.
%The nodes and the edges together form a network.
% The nodes 
% %(\textit{i.e.} robots) 
% are labeled according to their status of whether the robots are `informed’ by the information propagated through the swarm.
The flow of the nodes from one compartment to another
%(\textit{i.e.} uninformed robots get `informed’ with the information.)
occurs when active links exist between robots that are
`informed’ and `non-informed’ %the information 
(\textit{i.e.,} 
only `informed' robots %carrying the information 
can transmit the information to `non-informed' ones).%robots that are not yet informed). 

%\thales{However, epidemic models often operate under the assumption that the networks are static and deterministic, so that the flow between compartments can be analyzed with ordinary differential equations \cite{}. In reality, neither human contacts nor robots working in communication challenging environments can guarantee such consistent connectivity in large-scale networks. The classic compartmental-models-based control strategies for such networks also focus on the statistical behavior of the nodes, ignoring the topology of the networks. The analysis and solutions are, therefore, generally valid with giant node numbers reaching thermodynamic limits \cite{}. } 

However, compartmental models often focus on the statistical behavior of the nodes, overlooking the capabilities of the \textit{individual-level} decision making that may impact the network's performance.
%\xyucomment{I commented out this paragraph and edit it as a couple of bridge sentences. }
Recent works in robotics have noticed and embraced the potential of controlling individual robots so that the performance of the whole network (\textit{i.e.,} the transmission rates, if a propagation model is considered,) can be impacted or improved. %swarm can jointly achieve the desired status. 
The topology of networks was analyzed in \cite{Preciado2014} and control strategies were proposed to identify 
and isolate the nodes that have a higher impact on
the propagation process. The design of {\it time-varying} networks where robots leverage their mobility to move into each other's communication ranges to form and maintain temporal communication links has been addressed in  \cite{hollinger2012multirobot,khodayi2019distributed,yu2020synthesis}. Information can be propagated in such networks by relying on time-respect routing paths formed by a sequence chronologically ordered temporal links \cite{yu2020synthesis}. 
Nonetheless, such approaches still require thorough and perfect knowledge of the network's topology and the robots' ability to maintain active links and requires robots to execute their motion plans precisely both in time and space. As such, the resulting networks lack robustness to uncertainties in link formation timings that may result from errors in motion plan execution due to localization errors and/or actuation uncertainties.

%\thales{Stochastic models, such as networks based on percolation theory concepts \cite{}, have also been introduced in works in epidemiology in recent years to carry out a more realistic analysis. Percolation theory assumes a growing collection of edges between nodes. When the addition of edges fits a stochastic process, the growing graphs yielded are modeled as Erd\"os–R\'enyi networks, and have been well studied in the field of random graphs \cite{}. Noticed that such stochastic models focus on a random \textit{existence} of edges on a graph, echoing the assumption of Canadian Traveler Problems, in which the graphs are partially visible, and the edges are randomly blocked \cite{}. }

%Real-world communication networks synthesized in robot 
%swarms face uncertainties causing stochastic performance
%of the networks.
%that can not always be directly addressed by existing graph theory tools. 
Existing stochastic graph models focus on random changes in a network's topology centered around random creation and removal of edges within a graph. For example, the Canadian Traveller Problem assumes partially visible graphs where edges randomly appear and disappear \cite{bar1991canadian,nikolova2008route}.
For time-varying networks,
\cite{knizhnik2022flow,shen2022topology} assumed that any temporal link
may suffer from deviations from
its scheduled timing due to the uncertain arrival time of one robot into another robot’s communication range.
%\xyucomment{I commented out the original paragraph and edit the following graph as shown in blue. }
%Consider robot swarms moving in challenging environments
%with network topology changing over time.
%Any temporal link that occurs and disappears in such a network may suffer from deviations in its scheduled timing \cite{ACC2022} due to the uncertain arrival time of one robot into another robot’s communication range \cite{ICRA 2022}.
Information routing or propagation planned for such networks may experience severe delays if subsequent links along a routing path appear out of chronological order in the presence of link formation uncertainties resulting from uncertainties in robot motions \cite{yu2020synthesis}.  These challenges are addressed in \cite{shen2022topology} for networks with periodically time-varying interconnection topologies.  Control strategies to `fix’ nodes with 
higher impact on the whole network’s performance were also proposed in \cite{shen2022topology} similar in spirit to those presented in \cite{Preciado2014}.

In the Canadian Traveler Problems and \cite{shen2022topology}, messages are assumed to be transmitted instantaneously between nodes whenever a link appears. As such, the resulting strategies are solely based on the \textit{existence} of an available link between a certain pair of robots at a certain time point. When the time to transmit a message is non-trivial, the question goes beyond whether the information can be transmitted or not and must consider the quality of the transmission (\textit{e.g.,} safe, fast, confident). Consider a pair of robots flying through a Poisson forest maintaining a communication link that requires line-of-sight (LOS). The randomly distributed obstacles may intermittently disrupt the LOS between the robots causing randomized delay in the completion of the information transmission task. In these situations, it becomes difficult to determine whether a robot \textit{is} informed or not at a given time point which then impacts the robots ability to plan subsequent actions.

In this paper, we aim to develop receding horizon control schemes that allows individual robots to re-direct their transmission resources 
(\textit{e.g.,} time, power) to different neighboring robots on the communication network with stochastic
links to guarantee an exponentially fast
convergence status in the information propagation. We model the completion of the message transmission across a link as a Poisson Point Process. The density parameter of this process is determined by the transmission resources invested by the nodes.  Each node carries limited transmission resources that is distributed among all the links connecting to it.
The completion of the transition of a 
message from one node to another is then modeled as a Markov process. 
All robots would then be able to update the distribution of their own transmission resources according to the neighboring current 
states and follow the control 
directions. Therefore, the control
strategy takes into account the 
transmission resource capabilities of the
robots and acts on them to fulfill
performance requirements.
Such a set of resources changes 
according to the application, for
example, in the Poisson forest we
might require that the robots stay close
together at the cost of decreasing the
covered area; in a situation in which
the robots need to move back and forth
to carry information ({\it e.g.,} 
surveillance
around buildings or in tunnels), we
might require the robots to increase their
frequency of visits among each other.

The paper is organized as follows. Sec.~\ref{sec:graph_theory} and Sec.~\ref{sec:SI_model} provides theoretical backgrounds of the graph structure and the propagation model. The problem is formally stated at the end of Sec.~\ref{ProbForm}. Sec.~\ref{section3} introduces our approach of developing the receding horizon control scheme for the stochastic network. Sec.~\ref{simulation} validates our proposed control schemes with numerical examples. Sec.~\ref{conclusion} concludes the paper and proposes future directions.

\section{Background and Problem Formulation}
\label{ProbForm}

%In this section, we formally define 
%the problem tackled in this paper.
%Section \ref{sec:graph_theory} introduces the methodology
%applied to describe the network topology.
%In Section \ref{sec:SI_model} the agent's states and the state of the
%network are precisely defined.
%Finally,
%Section \ref{sec:control_statement} provides
%details on the
%control problem studied.

\subsection{Graph Theory}
\label{sec:graph_theory}
%We are interested in understanding the
%effects of stochastic
%communication topologies on the broadcast of information
%on networked systems.
We represent the communication network as a graph $\GG(\VV,\EE)$,
in which $\VV=\{1,...,n\}$ is the node set and
$\EE\subset \VV \times \VV$ the edge set.
Each element of the edge set represents a
\textit{directed} link from $i$ to $j$ 
if $(i,j)\in\EE$.
The set constituted by the nodes that have the
$i$th node as a child node is denoted by
$\NN_i=\{j\in\VV : (j,i)\in\EE\}$ and it is called
neighborhood of the $i$th node.
In accordance to this nomenclature, we call
a neighbor of a node $i$ a node $j$
in $\NN_i$.
The adjacency matrix $\A=[a_{ij}]$ associated with
the graph $\GG$ is an object that
maps the edge set into a matrix as,
\begin{align}
\label{eq:adjacency}
a_{ij}=\left\{
\begin{array}{ll}
0, & \text{if } i=j \text{ or } (j,i)\nexists \EE,  \\
\omega_{ij}(t), ~& \text{if } (j,i)\in \EE,
\end{array}
\right.
\end{align}
where $\omega_{ij}(t)>0$ is a positive number
that maps the transmission rate of information from node
$j$ to node $i$, which we will examine in more detail 
below.

\subsection{Information Propagation Model}
\label{sec:SI_model}

We study the stochastic broadcast of
information in a networked system represented
by a graph $\GG(\VV,\EE)$, in which the edges represent
communication channels and the nodes represent the robots.
We assume that each robot has limited communication 
capabilities,
in these scenarios the systems
may have to operate with
intermittent network connectivity due to
wireless signal attenuation, resulting from
obstacle occlusions and other
environmental factors \cite{Saddler2019}.
%We model
%such intermittent network connectivity 
%through the communication graph as 
%stochastic transmissions among nodes. 
We borrow an epidemiological model
from 
Van Mieghem and Cator \cite{Mieghem2012}, 
who defined an {\it exact} $2^n$-${\text{state}}$
joint Markov chain
to study the stochastic process of propagation of diseases 
using a Susceptible-Infected (SI) representation.
We investigate a similar
model to represent the broadcast of information.
Naturally, using $2^n$ states
to represent the broadcast of information can quickly
lead to computational infeasibility. 
To circumvent this problem
we apply a moment-closure 
to the model \cite{Cator2012,Schnoerr2015,Watikins2018},
which approximates the representation of the joint
Markov chain using a lower-dimensional representation.
Afterward, we 
compare the performance of both representations.

In a SI model, each individual can be either
``susceptible'' or ``infected'' by a disease, and
a susceptible agent can only become infected if
there is at least one infected agent in its neighborhood.
We adopt this idea to model the transmission of
information in a robot networked system,
such that each agent
can be either \textit{informed} or 
\textit{non-informed} about a 
message to be transmitted over the network.
We assume that the transmission process between
any given pair of agents $(j,i)$ 
follows a stochastic jump process with
Poisson distribution and transmission rate
$\omega_{ij}(t)$ \cite{Oksendal2007}.
%A scheme for the SI model is illustrated in Figure
%\thales{[ADD FIGURE]}.  
This type of modeling can represent 
scenarios where the information is carried
by mobile robots with low-range 
and limited communication capabilities,
which usually leads to intermittent interaction between
agents,
such as in \cite{Saddler2019,Yu2021}. 
%\thales{[the next part should be on the introduction]}
%As an example, consider
%applications such
%as surveillance, drifters in the ocean, and in GPS denied
%areas.

To be able to affect the transmission statistics over
the whole network while
satisfying a team's performance
criterion, 
we assume that the robots have 
access to the states of its neighborhood
on predefined sampling times, nevertheless, each
node can carry the computation with that information.
Even though this assumption can be conservative in 
general, there are scenarios in which it is possible to
have a higher range, but low transmission rate, to
perform general and simple coordination,
while agents need to
perform local higher demanding data transactions,
for example on data fusion tasks.
Also, estimates techniques might
help to alleviate such an assumption.

%\begin{rem}[Parameter and State Uncertainty]
%\thales{In the case we don't know those 
%parameters exactly,
%the way to go is to sample from the distribution and
%then infer other values.}
%\end{rem}

\begin{figure}[h]
\centering
\includegraphics[scale=.25]{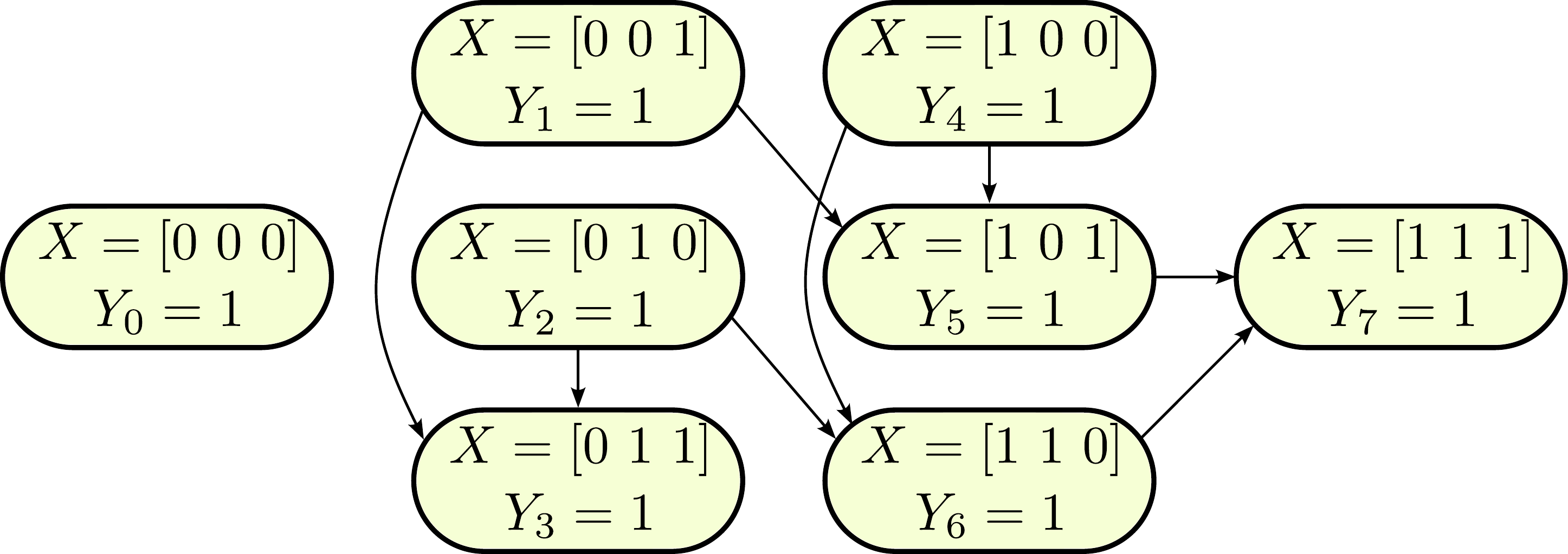}
\caption{{ Representation of all possible states
of a network with three agents. The arrows represent possible
switching between states. }
}
\label{fig:1}      
\end{figure}

We model the current state of a robot 
as an indicator 
random variable $X_i(t)\in \{0,1\}$, such that
$X_i(t)=1$ if the $i$th robot 
is informed at time $t$,
and $X_i(t)=0$ otherwise.
Note that $X_i(t)$ is a Bernoulli 
random variable, therefore
the evolution of the network expectation of
broadcast of information 
can be determined by
the evolution of the probability 
$\mathrm{Pr}\{X_i(t) = 1 : X_j(0) = 1\}$
for any pair $(i,j)$,
over time $t>0$.
The joint state of the network maps the set of agents
that hold the
information in a given instant.
Since $X_i(t)\in\{0,1\}$ for each $i\in\VV$,
it is clear that the joint states of the network
evolve in a $2^n$ state-space. 
Inspired by \cite{Mieghem2012,Sahneh2013} in epidemiological studies,
we define a
time-dependent state-space vector for the
whole network as
\begin{align*}
Y(t) ={}&[Y_0(t) ~ Y_1(t) ~ \cdots~Y_{2^{n-1}}]',
\text{ with}
\\
Y_i(t) ={}&\left\{
\begin{array}{ll}
1, & i = \sum_{k=1}^n X_k2^{k-1}, \\
0, & \text{otherwise,}
\end{array}
\right.
\end{align*}
for all $X_k\in\{0,1\}$. 
Observe that $Y(t)\in\real^{2^n}$ corresponds to
a binary-valued vector that maps
the current state of each node into a 
specific element $Y_i(t)$ ({\it i.e.,}
we have a bijective map between each possible state
of the network and the vector $Y(t)$). Fig.
\ref{fig:1} illustrates the relation of 
all possible states of the network and $Y(t)$
for a network with three nodes.
In addition, note that the elements of $Y(t)$
also are Bernoulli random variables, hence 
$\mathrm{E}[Y_i(t)]=\mathrm{Pr}\{Y_i(t)=1\}$.

One could try to model the states of the
whole network directly as $X(t)=[X_1(t)~X_2(t)~\cdots~X_n(t)]$,
since that would reduce the dimension of the 
representation from $\real^{2^n}$ to $\real^{n}$.
However, as shown in \cite{Sahneh2013},
the evolution of 
$\mathrm{E}[X(t)]$ represents only a {\it marginal 
probability distribution} of the states of the
network, and
this information is not enough to describe
the evolution of 
$\mathrm{Pr}\{X_i(t) = 1 : X_j(0) = 1\}$ for
any $t>0$.
Therefore, if we are interested in computing the expectation
of future states given a current distribution, using $X(t)$
alone
is not enough.
Alternative methods to reduce the dimension of the 
state-space include
mean-field type approximations and 
moment-closure methods (please, see \cite{Sahneh2013}
and references therein).
Typically those methods are valid under the
assumption of large numbers and 
have reduced precision about the underlying stochastic
process.
Indeed, as pointed out in \cite{Watikins2018},
in general, mean-field approximations do not
consistently provide
either an upper- nor a lower-bound for general
spreading processes.

Since $Y(t)$ is an exact characterization
of the states of the network,
an expression for
$\mathrm{Pr}\{Y_i(t)=1:Y(0)\}$
can be obtained using a
continuous-time Markov chain with
$2^n$ states, determined by the following
infinitesimal generator $Q(t)=[q_{ij}(t)] \in
\real^{2^n \times 2^n}$ 
\cite{Mieghem2012},
\begin{align}
\label{eq:transition_matrix}
q_{ij}(t) = \left\{
\begin{array}{ll}
\sum_{k=1}^n \omega_{mk}{X_k},
~& \text{for } m = 1,...,n, 
\text{  } X_m=1 \text{ if } 
i = j - 2^{m-1}, \\
-\sum_{k=1}^n q_{kj}, & \text{if } i=j,\\
0, & \text{otherwise,}
\end{array}
\right.
\end{align}
for all $X_k\in\{0,1\}$,
in which the corresponding Markov chain evolves according
the following Kolmogorov foward equation
\begin{align}
\label{eq:markov_2n}
\frac{dv(t)'}{dt}=v(t)'Q(t),
\end{align}
where the state vector is
defined as $v(t)=[v_0(t) ~\cdots~ v_{2^{n-1}}]'$ with elements
$v_i(t) = \mathrm{Pr}\{Y_i(t)=1\}$ and 
satisfies $\sum_{i}v_i=1$ for all $t>0$.

In this paper, we are interested in enforcing
a desired evolution for the propagation of
information on the network ({\it i.e.},
$\mathrm{Pr}\{X_i(t) \rightarrow 1 : X_j(0) = 1\}$)
exponentially fast,
by closing the loop of the network with
a feedback control strategy,
which chooses the control action
from a finite set of possible actions and minimizes a cost
of available resources. Specifically,
the problem investigated can be stated as:

\bigskip
\noindent \textbf{Problem 1.} 
Given a network with stochastic propagation of information
represented by the topology in \eqref{eq:adjacency},
define a feedback control that
actuates on the transmission rates $\omega_{ij}(t)$
such that
the evolution of the probability of
all nodes
in the network becoming informed nodes
follows
the desired convergence criteria.

\label{sec:control_statement}

\section{Methodology}
%\subsection{Formation of the communication links}
\label{section3}

\subsection{Control Strategy}
\label{sec:control_strategy}
To focus on the transmission of information over the
network and understand the ability to 
shape the broadcast of information, we make the following
assumption:
\begin{assumption}
The communication graph has a directed
spanning tree with a root corresponding
to the robot that is the origin
of the information to be transmitted.
\end{assumption}
This assumption implies that each edge
of the spanning tree has a positive transmission 
rate (\textit{i.e.,} the information is propagated with 
some
positive baseline transmission).
In this scenario, it is clear that the
broadcast of information will eventually reach 
every robot on the network with probability $1$.
%\thales{Explain how this scenario can be similar
%to other works, gyro like, Xi work, task allocation.}
Hence, the main objective of 
our control problem can be viewed as an attempt
to shape the statistics of the broadcast of 
information on the network to satisfy a given
convergence criterion.
It is worth noticing that
most works on contact processes are concerned
with regular lattices or mean-field models \cite{Philip2018,Preciado2014}.
Also, usually, the desired convergence rate is 
performed through static strategies, targeting
central nodes, and using heuristics \cite{Preciado2014,shen2022topology}.
With our control strategy, the network can
adapt to the stochastic realization of the propagation
of information and only expend resources to actuate during
realizations that might interfere with the team's desired
behavior.

With the broadcast model in \eqref{eq:markov_2n},
we control the transmission rates $\omega_{ij}(t)$
by applying allowable actions based on the current
state of the network and on predicted future states,
sampled on predefined times, following
a sampled-data predictive controller \cite{Pantelis2014}.
The controller we study in this paper
is defined by the following optimization problem:
\begin{align}
\label{eq:pred_ctrl}
& \min_{{\omega_{ij}}\in~U} C(X_i(t))
\\
&~s.t.~ 
\big(1-E\big[X_i(t+\Delta t):X(t)\big]\big)
-\big(1-X_i(t)\big)e^{-r\Delta t}\leq 0,
\nonumber
\end{align}
where $E\big[X_i(t+\Delta t):X(t)\big]$ is the
prediction stage which is 
computed using \eqref{eq:markov_2n}
(later we utilize a moment-closure to compute its
upper-bounds),
$C(X_i(t))$ is the cost of applying the
control action ${\omega_{ij}}$ on the transmission
between nodes $i$ and $j$, $\Delta t>0$ defines a
window of prediction from the current instant
$t$, the positive constant $r$ defines a
desired exponential convergence rate, and $U$ represents
the set of admissible control actions.
In practice, the set of admissible control
actions should represent the operations that robots 
can perform to adjust their
transmission rate among neighbors.
For example, if the team of
robots operate in
geographically disjoint locations and travel back
and forward regularly to neighbors' sites and carry
information with them,
one could define an increase in the transmission
rate as an increase in the frequency the agents should
visit each other. Another example is in the
case of periodic rendezvous, in which 
admissible actions could map the range of possible
changes in the agent's period.

We draw attention to the fact
that ensuring the constraints in 
\eqref{eq:pred_ctrl} plays a central role in guaranteeing
the fulfillment of the desired
performance. We notice that for the general case
it might be difficult to guarantee 
feasibility of problem
\eqref{eq:pred_ctrl}. Therefore, in this
work we {\it assume} that there is an auxiliary
control law that provides feasibility and that the 
controller has access to it. 
This assumption is not uncommon in the nonlinear
model predictive literature (see \cite{Ellis2014} and references
therein), typically such an auxiliary controller is 
provided {\it ad hoc}.
Nonetheless, in the problems we are interested in,
we notice that in several 
applications
it is possible to find feasible, but high-priced,
solutions for the problem. For example, in tasks where 
robots carry the message from site to site intermittently
we might satisfy the constraints by bringing all robots
closer to each other; in the case of robots performing
tasks on cyclic paths
we could make all robots cover overlaying paths.
Hence, even though it is necessary to provide an
auxiliary control strategy according to each task, we notice
that in the problem of broadcasting information simply
defining a feasible strategy is not necessarily
interesting. Therefore, we seek to apply
admissible control actions that do not lead to
trivial connectivity on the network and preserve
a satisfactory performance.

\subsection{Expected Time to Broadcast Information}
The feasibility of the optimization 
in \eqref{eq:pred_ctrl}
guarantees that the information is transmitted 
throughout the
network exponentially fast.
However, a remaining open question
is how can we infer
how long it will take
until all robots in the networks
receive the message of interest.
In other words, what is the hitting time for
the stochastic process to reach the stationary
state.
In this section, we analyze this question and provide
the expectation for the propagation time.

Given that we have the {\it exact}
representation of
the Markov chain that describes the
underlying stochastic
process we could, in principle,
compute the hitting times directly. However, since
we solve the receding optimization problem 
\eqref{eq:pred_ctrl}, our Markov chain is
non-homogeneous and its transition rates
might
change according to each realization.
Therefore, we take advantage of the
constraint in 
problem \eqref{eq:pred_ctrl}
relying on the assumption 
previously discussed that
a feasible solution can always be found and
compute the expected time to broadcast
information based on Watkins {\it et al.}
(Please, note that the method shown below
is similar to Theorem 5 in \cite{Watkins2020}.)

Summing the constraint in \eqref{eq:pred_ctrl}
over the $n$ nodes gives,
\begin{align*}
    n-E\big[\Scal(\boldsymbol{X}(t+\Delta t))\big]
    \leq (n-\Scal(\boldsymbol{X}(t))e^{-r\Delta t},
\end{align*}
where $\Scal(\boldsymbol{X}(t))=\sum_i^n X_i(t)$
is the number of informed robots 
at time $t$. 
The feasibility of the problem ensures that
\begin{align}
    n-E\big[\Scal(\boldsymbol{X}(t))\big]
    \leq (n-\Scal(\boldsymbol{X}(0))e^{-r t}.
    \label{eq:identity_from_constraint}
\end{align}
Define the broadcast time as 
$t_b=\inf\{t\geq 0:\Scal(\boldsymbol{X}(t))=n\}$,
then
its expectation can be computed as
\begin{align}
    E[t_b:\boldsymbol{X}(0)]
    =\int^{\infty}_0 1-{\rm Pr}(t_b\leq \sigma)d\sigma.
    \label{eq:expt_tb}
\end{align}
Note that ${\rm Pr}\{t_b\leq t\}={\rm Pr}\{\Scal(\boldsymbol{X}(t))=n\}$, therefore the
equivalence
$1-{\rm Pr}\{t_b\leq t\}
={\rm Pr}\{\Scal(\boldsymbol{X}(t))<n\}$ holds,
where ${\rm Pr}\{\Scal(\boldsymbol{X}(t))<n\}$ is
the probability of having any robot 
non-informed at
time $t$. We use this equality and
\eqref{eq:identity_from_constraint}
to
compute the expectation time for
the broadcast of information.
Note that the right-hand side of 
\eqref{eq:identity_from_constraint}
bounds the expected number of the existence
of non-informed robots, therefore
\begin{align*}
    {\rm Pr}\big\{\Scal(\boldsymbol{X}(t))<n
    :(\boldsymbol{X}(0))\big\}\leq
    \min \big\{1,(n-\Scal(\boldsymbol{X}(0))e^{-r t}\big\},
\end{align*}
hence, \eqref{eq:expt_tb} yields
\begin{align*}
    E[t_b:\boldsymbol{X}(0)]
    \leq & {~}
    \int^{\infty}_0 
    \min \big\{1,(n-\Scal(\boldsymbol{X}(0))
    e^{-r \lfloor \frac{\sigma}{\Delta t}
    \rfloor\Delta t}\big\}
    d\sigma
    \nonumber
    \\
    \leq & {~}
    \tau_1 + 
    \sum_{i=\frac{\tau_1}{\Delta t}}^\infty
    (n-\Scal(\boldsymbol{X}(0))\Delta t
    e^{-r \lfloor \frac{\sigma}{\Delta t}
    \rfloor\Delta t}
\end{align*}
\begin{align}
    \leq & {~}
    \tau_1 + 
    \frac{e^{-r\tau_1}}{1-e^{-r\Delta t}}
    (n-\Scal(\boldsymbol{X}(0))\Delta t,
\end{align}
with $\tau_1=\inf\{t\geq
0:(n-\Scal(\boldsymbol{X}(t)))e^{-r\lfloor \frac{t}{\Delta t}
    \rfloor\Delta t}\leq 1\}$.

\subsection{Robust Moment Closure}
\label{sec:robust_moment_closure}
In this section we consider a moment closure technique to
solve Problem $1$ with a computational amenable
method. We reiterate that, while \eqref{eq:markov_2n} 
is a theoretically right representation for our problem, it is not
computationally tractable for relatively medium-size networks due to
the necessity to use $2^n$ states to capture the evolution of the
probabilities in the network.
The technique considered in this section was introduced 
in \cite{Watikins2018} for the problem of continuous-time epidemics.
The approach is based on the Fr\'echet inequalities,
defined as
\begin{align}
    \FF_{\rm lwr}({\rm Pr}\{A\},{\rm Pr}\{B\})=&{~}\max \{0,{\rm Pr}\{A\}+{\rm Pr}\{B\}-1\},
    \\
    \FF_{\rm upr}({\rm Pr}\{A\},{\rm Pr}\{B\})
    =&{~}\min \{{\rm Pr}\{A\},{\rm Pr}\{B\}\},
\end{align}
which are lower- and upper-bounds for the joint probability ${\rm Pr}\{A,B\}$, respectively.
Its main
advantage over mean-field approximations
is that it gives meaningful bounds for the evolution of the
real mean of the underlying stochastic process (see \cite{Watikins2018} for a detailed discussion).

Initially, we need to reformulate our representation to
consider the marginal 
probabilities of each robot receiving 
the information, since
we are interested in drawing bounds on their evolution.
For our problem,
the marginal probabilities can be written as the expectation of
a robot becoming informed as a function of its neighbors as,
\begin{align}
    \frac{dE[X_i(t)]}{dt}=\sum_{j\in \NN_i}E[\bar X_i(t)X_j(t)]\omega_{ij}(t),
    \label{eq:marginal_de}
\end{align}
for all $i\in \VV$, where $\bar X_i(t)$ is
 the complement of $X_i(t)$ ({\it i.e.,} 
 $\bar X_i(t)=1$ if $X_i(t)=0$).
 This equation maps an infinitesimal
probability of $X_i$ receiving the information
given that some neighbors $X_j$, $j\in\NN_i$, have the message.
Notice that \eqref{eq:marginal_de} is not closed, therefore
we cannot solve it for $E[X_i(t)]$ without equations for 
$E[\bar X_i(t)X_j(t)]$.
Central to the result in \cite{Watikins2018}
is to bound the joint probabilities of the form
$E[\bar X_iX_j]$.
Next, we show 
the optimal approximations for the
expectations we employ,
which can be expressed as closed-form functions.
\begin{lemma}[Watkins {{\bf et al.,}} \cite{Watikins2018}]
Let $\underbar{x}(0)=\boldsymbol{X}(0) = \overline{x}(0)$,
define $\Scal(\boldsymbol{X})$ as the number of nodes
informed in ${\boldsymbol X}$. It holds
that
\begin{align}
    E[\Scal(\boldsymbol{X}(t)):\boldsymbol{X}(0)]
    \leq & {} \sum_{i\in \VV}\min \Big\{
    \overline{x}_i^{\rm inf}(t),
    1-\underbar{x}_i^{\rm non}(t)\Big\}, \text{ and}
    \\
    E[\Scal(\boldsymbol{X}(t)):\boldsymbol{X}(0)]
    \geq & {} \sum_{i\in \VV} \max \Big\{
    \underbar{x}_i^{\rm inf}(t),
    1-\overline{x}_i^{\rm non}(t)\Big\},
\end{align}
where the variables $\overline{x}^{\ell}_i(t)$ and
$\underbar{x}^{\ell}_i(t)$, for each
$\ell\in\mathcal{L}=\{{\rm inf},
{\rm non}\}$ are the solutions of the following 
ordinary differential equations,
\begin{align}
    \label{eq:momentcls_upper}
   \dot{\overline{x}}_i^\ell=&{}\sum_{j\in\NN_i} 
   \sum_{\ell'\in\mathcal{L}}\FF_{\rm upr}(\BB_{\overline{x}_i^{\ell}}
   (\overline{x}^{\ell'}_i),\overline{x}_j^\ell)\omega_{ij}(t),
   \\
    \label{eq:momentcls_lower}
   \dot{\underbar{x}}_i^\ell=&{}\sum_{j\in\NN_i} 
   \sum_{\ell'\in\mathcal{L}}\FF_{\rm lwr}(\underbar{x}^{\ell'}_i,\underbar{x}_j^\ell)\omega_{ij}(t),
\end{align}
and $\BB_{\overline{x}_i^{\ell}} (\overline{x}^{\ell'}_i)
=\min\big\{1-\overline{x}_i^{\ell}, \overline{x}^{\ell'}_i\}$
is the upper-bounding operator.
In addition, the following inclusion is always satisfied
\begin{align*}
    E[{X}^\ell_i(t):\boldsymbol{X}(0)]
    \in \big[
    \underbar{x}_i^{\ell}(t),
    \overline{x}_i^{\ell}(t)\big]\subseteq\big[0,1\big],
\end{align*}
for all $t\geq0$.
\label{lemma:moment_cls}
\end{lemma}
The result in Lemma \ref{lemma:moment_cls} allows us to
compute bounds for the evolution of the probabilities 
of the broadcast of information in the network using
$4n$ differential equations (equations \eqref{eq:momentcls_upper}
and \eqref{eq:momentcls_lower} for each informed
and non-informed node), instead of $2^n$.
This enables us to apply the receding horizon
strategy in equation \eqref{eq:pred_ctrl} in larger networks.

\section{Numerical Evaluation and 
Discussions}
\label{simulation}
In this section,
we illustrate the approach proposed in Section
\ref{section3} with two simulation results.
The first simulation was carried out on a 15 robots static 
topology. The topology of the network was generated using Erd\"{o}s-R\'{e}nyi model
with connection probability of 0.3. 
We assumed that each robot has an uniform total transmission rate $\mu$ to distribute to its connected neighbor.
% In other words, once a robot become informed, 
% it can transmit the message to
% its neighbors with different
% transmission rates $\omega_{ij}(t)$ such that 
% the sum its transmission resources 
% is equal to an uniform fixed value $\mu$.
% In this simulation, we set the total
% transmission rate $\mu$ to be 2.
In other words, an informed robot 
can become connected to
its neighbors according to different
transmission rates $\omega_{ij}(t)$
such that 
the sum of its transmission rates 
between neighbors
is limited by fixed value $\mu$.
This parameter, along with the connection probability of Erd\"{o}s-R\'{e}nyi model, has a big impact on the information propagation speed of the whole network. In this simulation, we set the total
transmission rate $\mu$ to be 2 for better comparison results.
% and $\omega(t)$ is randomly sampled so that  $\omega(t) \in (0,\mu)$,  $\sum_{j}^ {j\in\VV , (j,i)\in\EE} \omega_{ij} = \mu$.
% \thales{and we  each edge's transmission rate following this assumption. }
% \cthales{Not sure what do you mean.}
% \begin{align}
% \label{eq:pred_ctrl}
% & \min_{{\omega_{ij}}\in U} C(X(t))
% \\
% &~s.t.~ 
% n-E\big[\Scal(\boldsymbol{X}(t+\Delta t))\big] - 
%      (n-\Scal(\boldsymbol{X}(t))e^{-r\Delta t} \leq 0,
%     \\
% \big(1-E\big[X_i(t+\Delta t):X(t)\big]\big)
% -\big(1-X_i(t)\big)e^{-r\Delta t}\leq 0,
% \\
% & \sum_{j}^ {j\in\VV , (j,i)\in\EE} \omega_{ij} = \Omega
% \\
% & \omega_{ij} \in (0, \Omega)
% \nonumber
% \end{align}

\begin{figure}[h]
\centering
\begin{subfigure}[t]{0.49\textwidth}
\centering
\includegraphics[width=1\textwidth]{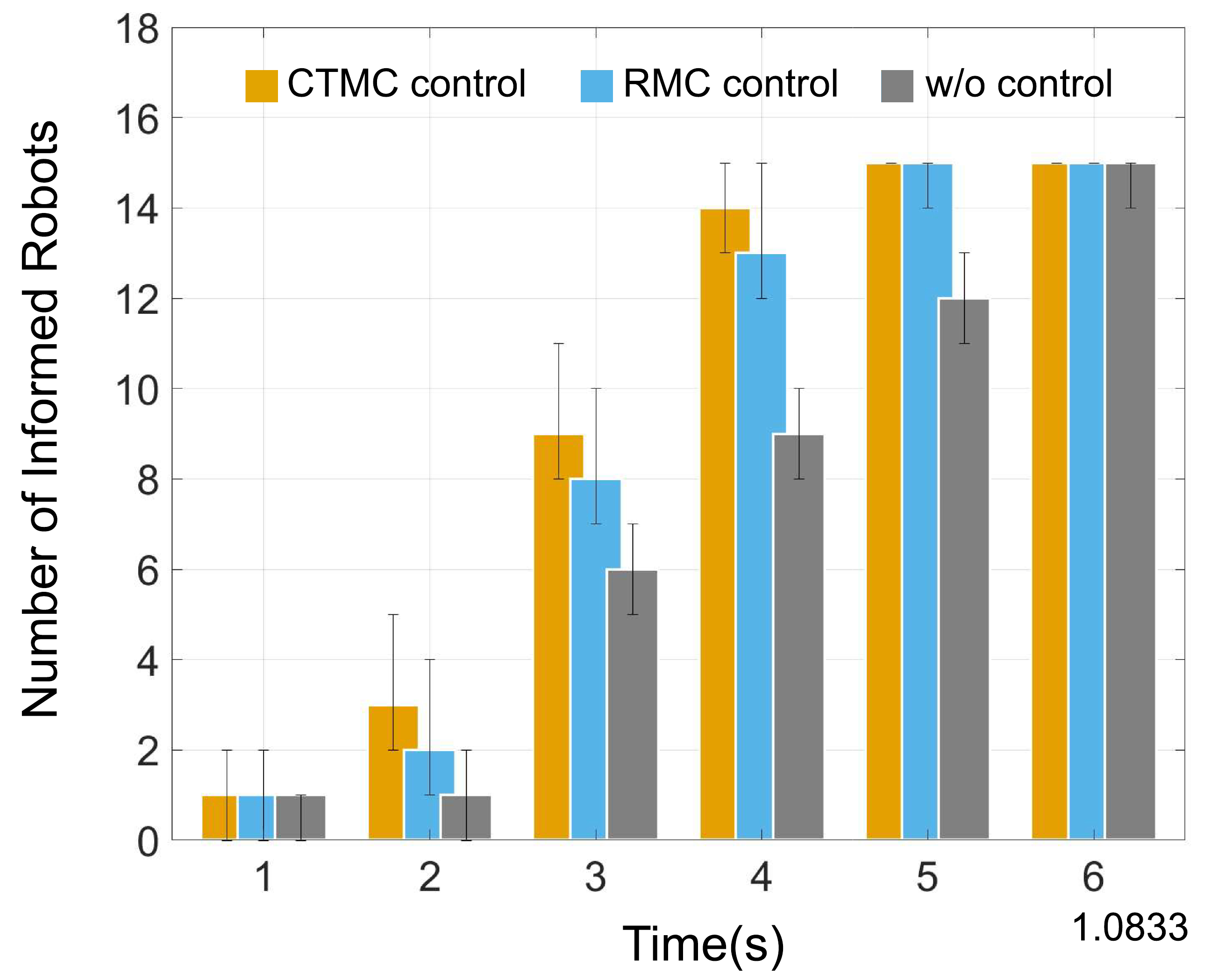}
\caption{}
\label{fig:15bar}
\end{subfigure}
\hfill
\begin{subfigure}[t]{0.49\textwidth}
\centering
\includegraphics[width=1\textwidth]{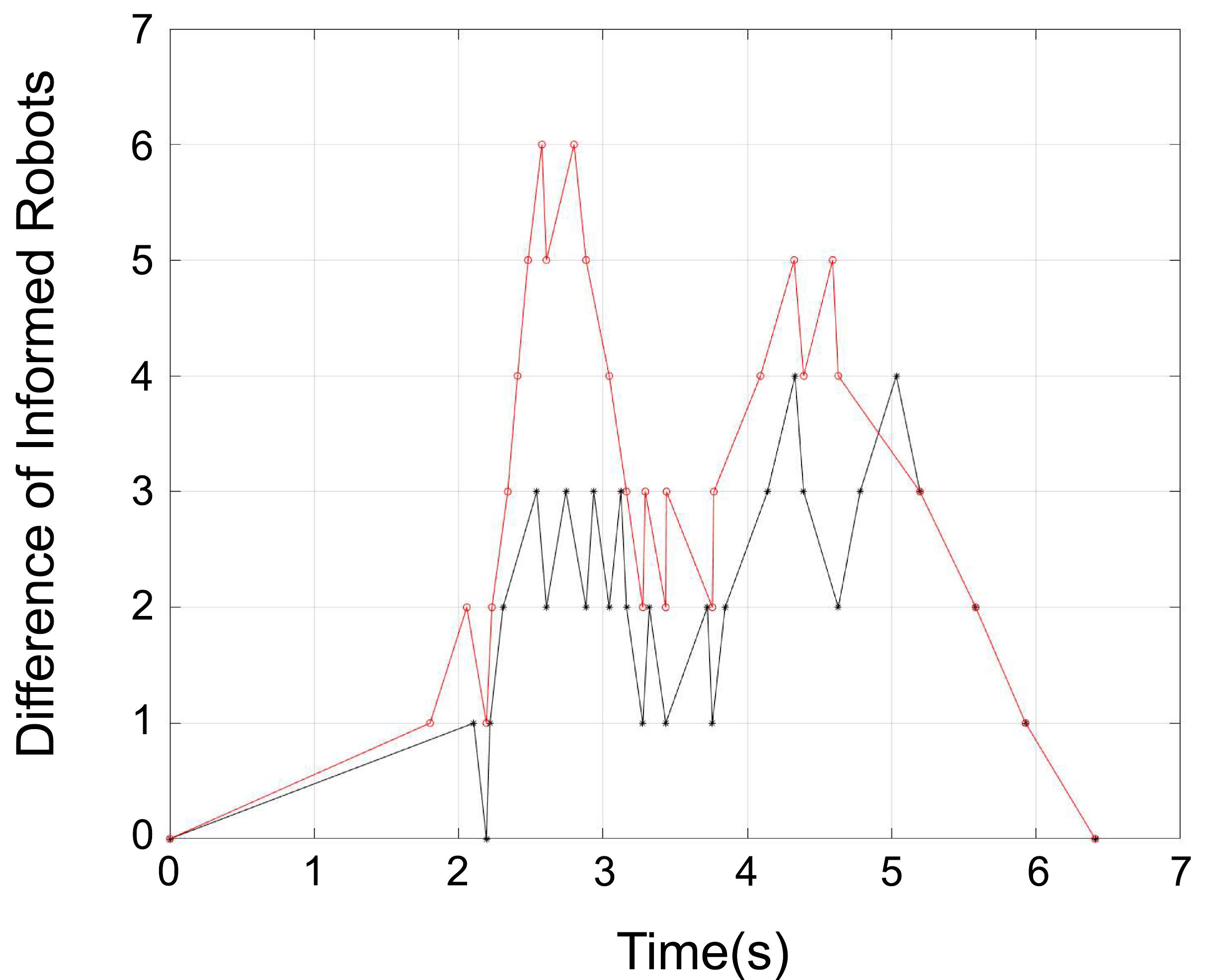}
\caption{}
\label{fig:15diff}
\end{subfigure}

\caption{{
Numerical results comparison of 15 nodes network’s broadcasting speed. 
In (a), the bar chart showcases the total number of robots that received information after the end of each 1.0833 seconds. The error bars represent the maximum and minimum informed robots in our 10 different data-set. In (b), the Line chart demonstrates the evolution of the difference in informed robots with respect to time. The red plot shows the difference between CTMC and the exact model, and the black plot shows the difference between RMC and the exact model.
} 
}
\label{fig:15_simulation_result}

\end{figure}

We applied the control strategy \eqref{eq:pred_ctrl} proposed in 
Section \ref{sec:control_strategy} with the
Con-tinuous-Time Markov Chain (CTMC) representation
and with the Robust Moment Closure (RMC), model as
described in Section \ref{sec:robust_moment_closure}.
We can choose from a myriad of cost functions that align with different design purposes.
Here we chose $C(X_i(t))=\sum_{j\in\NN_i}\omega_{ij}^2$. If we were not seeking an exponential propagation, this objective function, together with the constraints on the transmission
resource $\omega_{ij}(t) \in (0,\mu)$, and
$\sum_{j\in\NN_i}\omega_{ij} = \mu$ for
each $i\in \VV$, would push for an even distribution of every robot's resources between its neighbors. We chose $\Delta t=1$, 
and $r=0.22$ for both CTMC and RMC models, and simulated our proposed controller which guarantees an exponential propagation while 
minimizes the chosen cost.
%The control method was defined as 
%in \eqref{eq:pred_ctrl} with
%$C(X_i(t))=\sum_{j\in\NN_i}\omega_{ij}^2$, $\Delta t=1$, 
%and $r=0.22$ for both CTMC and RMC models.
%The set of admissible control actions $U$ 
%is defined by the constraints $\omega_{ij}(t) \in (0,\mu)$, and
%$\sum_{j\in\NN_i}\omega_{ij} = \mu$ for
%each $i\in \VV$.
The original exact model is also simulated
without any control method in the loop. 
We run all three models 10 times each, and we took their average as the performance results. The performance comparison result is shown in Fig. \ref{fig:15_simulation_result} and the numerical result at different stages of the propagation process is in Table.~\ref{tab:results}.

 From Fig.\ref{fig:15bar},
the CTMC model with control outperforms two other models,
the RMC model with control is slightly slower than
the CTMC, but still faster than the original model without control.
The CTMC model was in average  39.6$\%$ faster than the open-loop model to informed all robots, and RMC model was 27.3$\%$ faster than the open-loop model completion time.
% \cthales{Instead of the \% of the total time, can you compute how \% faster it was? It's just the complement and is simpler for the reader}
% The CTMC model took 71.62$\%$ of the time comparing with the open-loop model completion time to informed all robots, and RMC model finished using 78.52$\%$ of the open-loop model completion time.
% \thales{The CTMC model informed all robots in 4.5902 seconds, RMC model finished in 5.0324 seconds, and epidemic model in 6.4092 seconds.}\cthales{Put those as percentage in relation to wo/ controls.}
Notice that the two patterns in black
and in red lines shown in Fig. \ref{fig:15diff} are similar--the biggest difference number between CTMC and RMC is 3 robots which occurs around 2.5 seconds.
%\shen{here the two - is not shown on the paper}

% \begin{figure}[h]
% % Use the relevant command for your figure-insertion program
% % to insert the figure file.
% % For example, with the graphicx style use
% \centering
% \begin{subfigure}[t]{0.49\textwidth}
% \centering
% \includegraphics[width=1\textwidth]{figures/15nodes_final.pdf}
% \caption{}
% \label{fig:15nodes}
% \end{subfigure}
% \hfill
% \begin{subfigure}[t]{0.49\textwidth}
% \centering
% \includegraphics[width=1\textwidth]{figures/100node_final.pdf}
% \caption{}
% \label{fig:100nodes}
% \end{subfigure}

\begin{figure}[h]
\centering
\begin{subfigure}[t]{0.49\textwidth}
\centering
\includegraphics[width=1\textwidth]{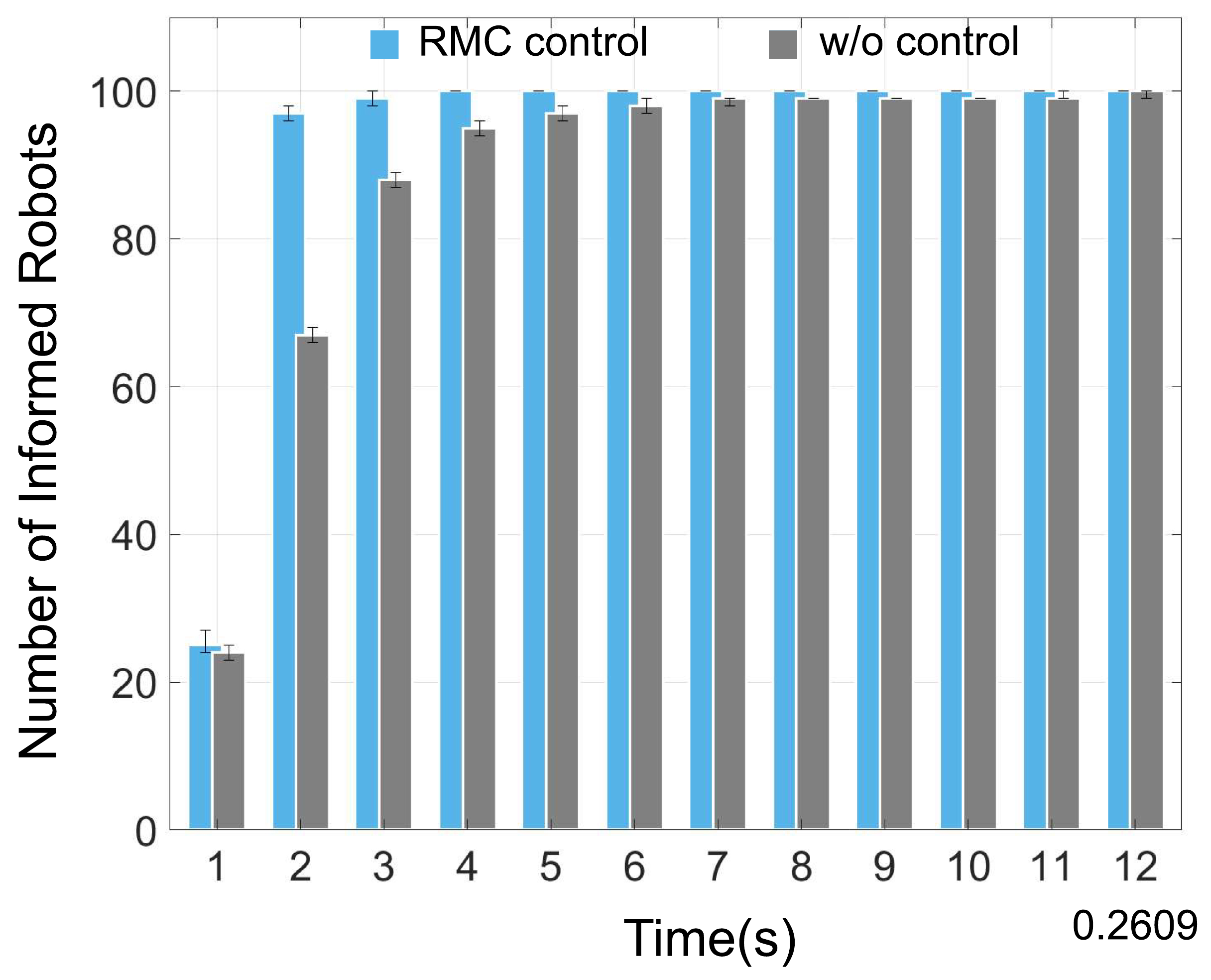}
\caption{}
\label{fig:100bar}
\end{subfigure}
\hfill
\begin{subfigure}[t]{0.49\textwidth}
\centering
\includegraphics[width=1\textwidth]{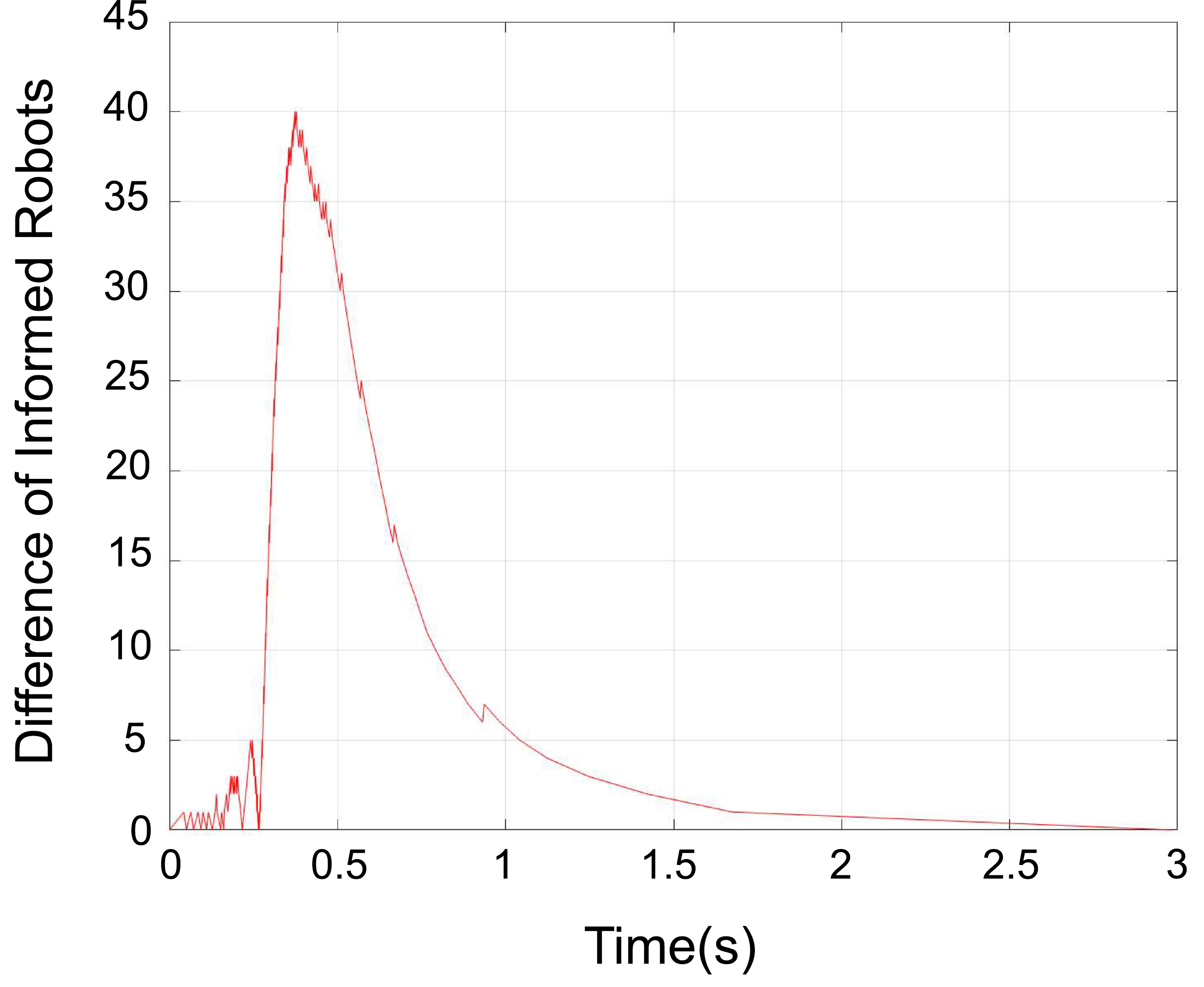}
\caption{}
\label{fig:100diff}
\end{subfigure}

\caption{{
Performance comparison of 100 nodes network’s broadcasting speed. 
In (a), the bar chart depicts the total number of informed robots after every 0.2609 seconds for both RMC and exact models. The error bars indicate the variation in informed robots of 10 different simulations.
In (b), the line plot represents 
the evolution of the difference
in informed robots between RMC and the open-loop method.
} 
}
\label{fig:100_simulation_result} 
\end{figure}

%Table \ref{tab:results} shows the average accumulated times that all models take to inform $20\%, 40\%, 60\%, 80\%$, and $100\%$ of the robots during the propagation process of different topology networks.

\begin{table}[h]
    \centering
    \caption{Average completion time at each propagation stage}
    \begin{tabular}{c|c|c | c | c | c | c}
    \toprule
    Topology & Model & $20\%$ & $40\%$ & $60\%$ &$80\%$& $100\%$ \\ \midrule 
    \multirow{3}{*}{\parbox{2cm}{\centering 15 nodes}}&
    CTMC &$\ 2.0571\ $ & $\ 2.4084\ $  &$\ 2.7997\ $ & $\ 3.7673\ $ & $\ 4.5902\ $ \\
    & RMC &$\ 2.2174\ $ & $\ 2.7436\ $  &$\ 3.3202\ $ & $\ 4.1394\ $ & $\ 5.0324\ $ \\
    &\ Open-Loop\ &$\ 2.6085\ $ & $\ 3.1619\ $  &$\ 3.7573\ $ & $\ 5.1964\ $ & $\ 6.4092\ $ \\ \midrule
    \multirow{2}{*}{\parbox{2cm}{\centering 100 nodes}}&
    RMC &$\ 0.2325\ $ & $\ 0.2849\ $  &$\ 0.3131\ $ & $\ 0.3525\ $ & $\ 0.9364\ $ \\
    &\ Open-Loop\ &$\ 0.2515\ $ & $\ 0.3343\ $  &$\ 0.4603\ $ & $\ 0.6428\ $ & $\ 2.9845\ $ \\ \bottomrule
    \end{tabular}
    \label{tab:results}
\end{table}
%

% \begin{figure}[h]
% % Use the relevant command for your figure-insertion program
% % to insert the figure file.
% % For example, with the graphicx style use
% \centering
% \begin{subfigure}[t]{0.49\textwidth}
% \centering
% \includegraphics[width=1\textwidth]{figures/15nodes_final.pdf}
% \caption{}
% \label{fig:15nodes}
% \end{subfigure}
% \hfill
% \begin{subfigure}[t]{0.49\textwidth}
% \centering
% \includegraphics[width=1\textwidth]{figures/100node_final.pdf}
% \caption{}
% \label{fig:100nodes}
% \end{subfigure}

% \caption{{
% Numerical results comparison of network’s broadcasting speed. The upper line chart demonstrates the evolution of the difference in the number of informed robots between our approaches and the open-loop approach with respect to time. The lower bar chart showcases the total number of robots that received information after the end of each time interval. (a) The red plot in the upper graph represents the difference between CTMC and the exact model, and the black plot represents the difference between RMC and the exact model. (b) The upper graph shows the number of informed robots’ difference between RMC and the open-loop method over time. The red bar in the lower graph is the open-loop model without control, and the blue bar is RMC with control.
% } 
% }
% \label{fig:simulation_result} 

% %
     
% \end{figure}

We conducted the second simulation in a
larger static topology network with 100 nodes.
The CTMC model is not appropriate in this setting as the state-space evolves in
$\real^{2^{100}\times2^{100}}$. The topology of the network is again generated using Erd\"{o}s-R\'{e}nyi model
with connection probability of 0.05,
and total transmission rate $\mu=4$.
We compared the performance between the open-loop
model and RMC model in broadcasting speed.
The control strategy is implemented
considering $\Delta t =0.1$ and 
exponential convergence rate
$r=2.8$. We run 10
different trials for each model
using random seeds and use their mean values as the numerical results.

% \begin{figure}[h]
% \sidecaption[t]
% % Use the relevant command for your figure-insertion program
% % to insert the figure file.
% % For example, with the graphicx style use
% \includegraphics[scale=.6]{figures/100node_comparision.pdf}
% %
% \caption{{\small Numerical result of network's broadcasting speed with Robust Moment Closure with control and the original epidemic control. The lower bar chart showcases the total number of robots have received information after each time interval. The upper plot demonstrates the evolution in number of informed robots difference using Robust Moment Closure model with control compared to epidemic model with respecting to time. \normalsize}
% }
% \label{fig:2method}      
% \end{figure}

Fig. \ref{fig:100_simulation_result} shows the performance in
information broadcasting for those two models.
Notice that at the beginning in Fig. \ref{fig:100bar}, before 0.3 seconds,
there was not much difference between the two
methods.  
This is because the constraint of the optimization problem was satisfied with the initial trivial solution and 
there was not many actions in $U$ that could further
improve the information propagation during
that period.
This might be why there is a fluctuating pattern at the beginning in Fig. \ref{fig:100diff}.
After that, the control strategy
starts to show its advantage,
and the average time to broadcast the information to all robots is 
%0.9364 
 0.93 seconds, which is in average 218.7$\%$ faster than the open-loop network, which has
 an average of 2.98 seconds.
%  The exact model finishes in 2.98 seconds,
%  \thales{that is \{\} faster
%  than the open loop network. \{add \% of how fast this is\}}.
 %2.9845 seconds.

Both numerical examples show a significantly better performance of our proposed methods in the scenarios assuming a random node is propagating information to the rest of the team. Notice that such scenarios are not usually seen in real-world applications. A faster propagation indicates a more efficiently connected network, providing a tighter upper bound of the information transmission time between any pair of nodes in the network. Therefore the propagation models were chosen to demonstrate a more efficient transmission is expected for any shortest path connecting two nodes in the network. A limitation of this propagation model is that all shortest paths are considered equally important, while in real-world applications, it makes sense that certain paths are with higher priorities due to the actual variation of demands. Our method can be easily tailored to fit different objective functions.

\section{Conclusions and Future Work}
\label{conclusion}
We have investigated a control structure for shaping the transmission rates on networked systems
to increase the connectivity in networks, the broadcast of information can be modeled as stochastic jump processes. As a first result, we have shown that by using an epidemiological model to represent the transmission of information among robots 
we can describe the probability of future transmissions based on the current joint state of the network and its topological structure. Then, based on this finding, we examined the applicability of a receding control strategy to account for possible drifts in performance on the current realization and, subsequently, actuate on transmission rates whenever necessary. This approach provides efficient adjustments to the network. Finally, numerical experiments were implemented, illustrating the effectiveness of the method. Possible future work includes the extension of the strategy for distributed optimization. It is also essential to connect the robot dynamics with the changing transition rates. One possible approach to bridging those two models is through maps defined {\it a priori} and then accounting for them on the admissible set of actions. Finally, our methodology only accounts for the expectation of the transmission of information. This parameter can be a weak decision parameter, an extension to consider the variance would be appealing.

\section*{Acknowledgements}
We gratefully acknowledge the support of ARL DCIST CRA W911NF-17-2-0181,
Office of Naval Research (ONR)
Award No. N00014-22-1-2157.

%\section{Appendix}
%\label{appendix}
%\input{sections/7_appendix.tex}

                                  % This command does not take effect until next page

%%%%%%%%%%%%%%%%%%%%%%%%%%%%%%%%%%%%%%%%%%%%%%%%%%%%%%%%%%%%%%%%%%%%%%%%%%%%%%%%

\bibliographystyle{spmpsci}
\bibliography{main.bbl}

\end{document}